\ifdefined\pdfminorversion
\fi

\documentclass[10pt,twocolumn,letterpaper]{article}

\usepackage[pagenumbers]{cvpr} 

\usepackage{placeins}
\usepackage{colortbl}
\definecolor{TableBest}{HTML}{FFC8C8}
\definecolor{TableSecond}{HTML}{FFE5C8}
\definecolor{TableThird}{HTML}{FFFFC8}
\newcommand{\bestscore}[1]{\cellcolor{TableBest}\textbf{#1}}
\newcommand{\secondscore}[1]{\cellcolor{TableSecond}#1}
\newcommand{\thirdscore}[1]{\cellcolor{TableThird}#1}

\definecolor{cvprblue}{rgb}{0.21,0.49,0.74}
\usepackage[pagebackref,breaklinks,colorlinks,allcolors=cvprblue]{hyperref}

\def\confName{CVPR}
\def\confYear{2027}

\title{ArgsWorld: Unifying 3DGS Scene Reconstruction, Generation and Extrapolation with Autoregressive Diffusion Transformer}

\author{
Weiming Chen\textsuperscript{1},
Shuaiyan Liu\textsuperscript{2},
Xinyu Shao\textsuperscript{1},
Zhewen Chen\textsuperscript{1},\\
Tian Guan\textsuperscript{1},
Yonghong He\textsuperscript{1,\(\dagger\)},
Xitong Ling\textsuperscript{1,\(\dagger,\ddagger\)}\\[0.5em]
\small \textsuperscript{1}Tsinghua Shenzhen International Graduate School, Tsinghua University\\
\small \textsuperscript{2}Harbin Institute of Technology
}

\begin{document}
\maketitle
\begingroup
\renewcommand{\thefootnote}{}
\footnotetext[0]{$^{\dagger}$Corresponding authors.\quad $^{\ddagger}$Project lead.}
\endgroup
\begin{abstract}
Autoregressive diffusion world models deliver strong video generation performance, yet they typically output only RGB frames rather than explicit 3D Gaussian Splatting (3DGS) scene representations, creating a gap between learned visual world dynamics and usable 3D assets for downstream robotics, simulation and rendering. To bridge this gap, we introduce \textbf{ArgsWorld}, an autoregressive diffusion model that also produces an explicit 3DGS scene. For each generation chunk, the diffusion transformer jointly processes clean historical image latents and noisy latents for new views. It keeps the historical image latents fixed and refreshes the scene features so that the Gaussian representation can be revised as new regions come into view. Our 3D Self-Forcing Distillation trains the model on its own generated histories using a mixture of single-view and multi-view data. Distribution matching supervises direct image latents and latents of Gaussian renderings at both generation and additional cameras. ArgsWorld unifies 3D scene reconstruction from posed images, scene generation from image or text prompts, and scene extrapolation across camera segments with disconnected endpoints. Experimental results demonstrate that ArgsWorld achieves state-of-the-art performance across multiple 3D scene modeling tasks. Project page: \url{https://m4a1tastegood.github.io/ArgsWorld/}
\end{abstract}
\suppressfloats[t]
\begin{figure}[t]
    \centering
    \includegraphics[pagebox=cropbox,trim=0 9bp 0 19bp,clip,width=\columnwidth]{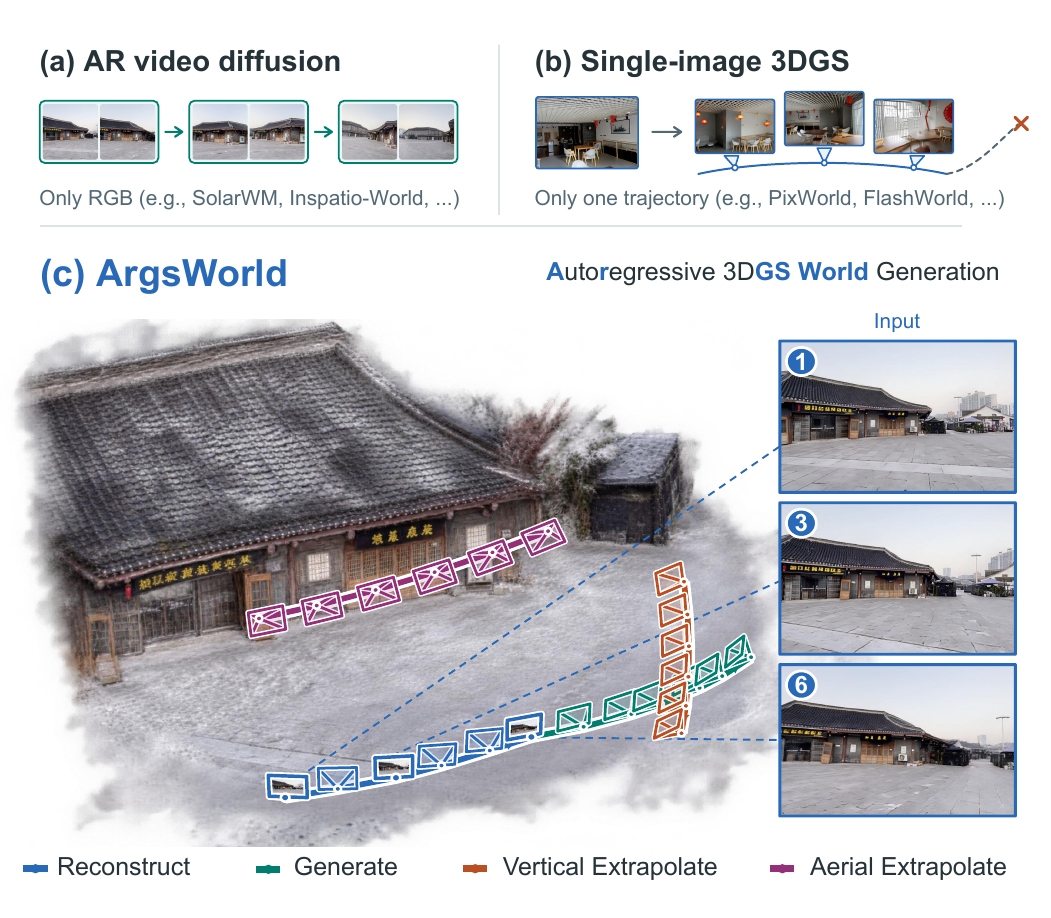}
    \caption{\textbf{ArgsWorld} is an \textbf{a}uto\textbf{r}egressive 3D\textbf{GS} \textbf{world} model that unifies 3D scene reconstruction, trajectory-guided view generation, and open-ended scene extrapolation. It reconstructs a 3D Gaussian scene from posed views (blue), generates views along the trajectory (green), and extrapolates along vertical (orange) and aerial (purple) camera paths.}
    \label{fig:teaser}
\end{figure}    
\section{Introduction}
\label{sec:intro}

Video diffusion models provide strong priors for visual synthesis~\cite{wan2025wan}. Autoregressive generation extends these models beyond a single sampling window~\cite{huang2026self,rui2026biwm}, and camera conditioning lets users specify where to look next~\cite{rui2026biwm}. These models describe a scene through the RGB frames they generate along a trajectory. An explicit scene representation would also allow users to render the same world from additional cameras and reuse it as a 3D asset.

\begin{figure*}[t]
    \centering
    \includegraphics[pagebox=cropbox,width=\textwidth]{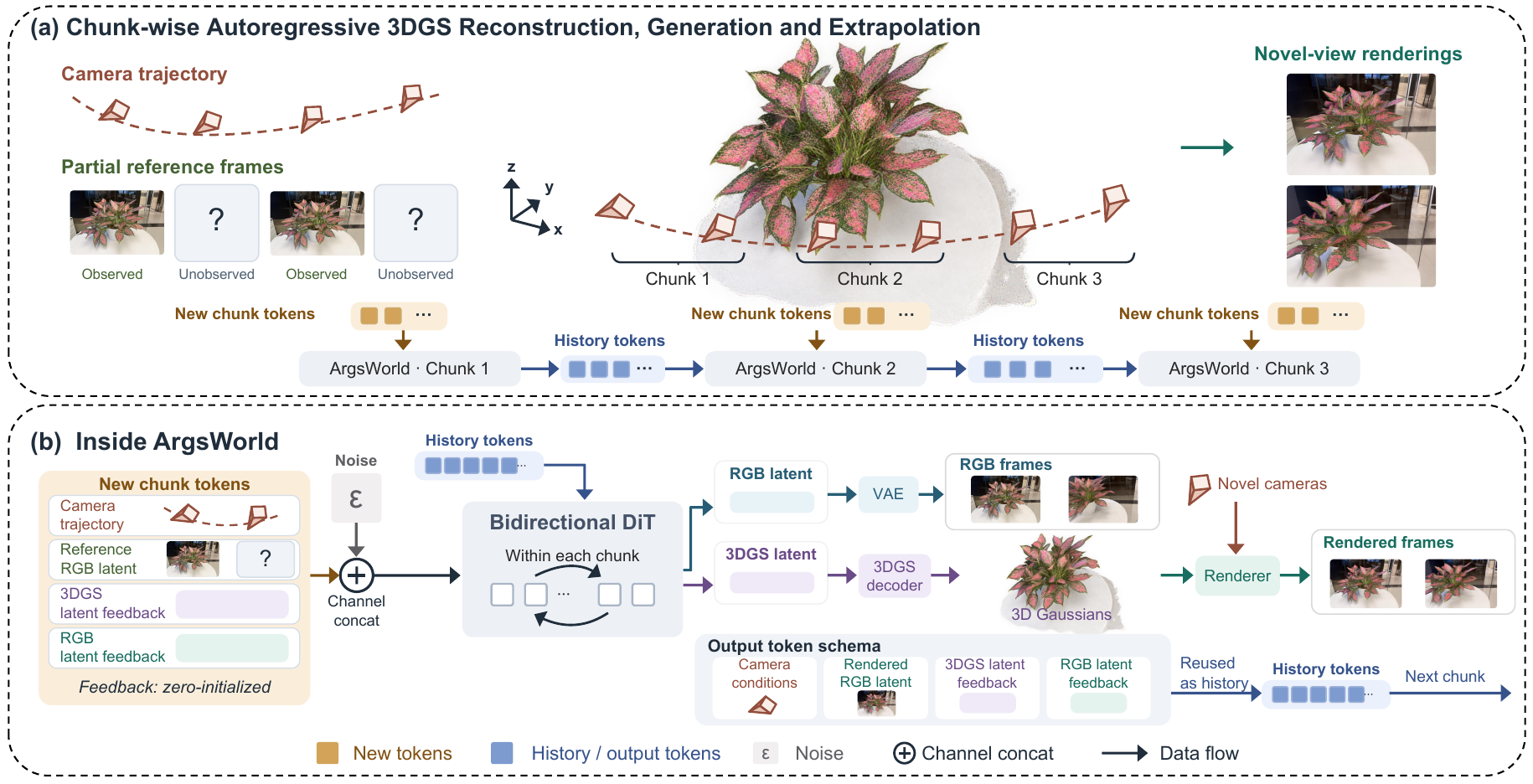}
    \caption{\textbf{ArgsWorld inference.}
    (a) We process a camera trajectory in successive chunks, using available reference images to condition the first chunk. Multiple input images enable scene reconstruction, while a single input image supports open-ended scene generation.
    (b) Each DiT call attends jointly to the current views and their clean history, predicting new RGB latents and refreshing features for Gaussian decoding across all views processed so far. 3DGS latents and DiT features feed the next denoising step within a chunk. Across chunks, the model carries forward its final direct RGB latents and 3DGS latents.}
    \label{fig:inference}
\end{figure*}

3D Gaussian Splatting (3DGS) represents a scene with explicit primitives and supports differentiable rendering~\cite{kerbl2023gaussians}. Feed-forward methods such as pixelSplat~\cite{charatan2024pixelsplat}, MVSplat~\cite{chen2024mvsplat}, and AnySplat~\cite{jiang2025anysplat} recover Gaussian scenes from observed images. FlashWorld~\cite{flashworld2025} extends this capability to scene generation by coupling a video diffusion backbone with a Gaussian decoder, and PixWorld~\cite{gao2026pixworld} unifies reconstruction and generation in pixel space. Building on these approaches, we ask how a generative model can revise and extend an explicit Gaussian scene as new camera queries arrive.

Three challenges arise when generating a Gaussian scene over successive chunks of camera views. Plausible image predictions provide only an indirect constraint on the geometry needed to render them. As the trajectory reveals new regions, newly predicted geometry must agree with the already generated appearance in overlapping views. Finally, a model trained with ground-truth histories encounters a different distribution of contexts when it must condition on its own predictions at inference time~\cite{huang2026self}.

We introduce \textbf{ArgsWorld}, an \textbf{Autoregressive 3DGS World Model} for reconstruction, generation, and spatial extrapolation. In \cref{fig:teaser}, ArgsWorld reconstructs a 3D Gaussian scene from three posed images, generates views along the initial trajectory, and extends the scene along vertical and aerial camera paths. We build on the camera-conditioned, 3D-adapted Wan backbone of FlashWorld~\cite{wan2025wan,flashworld2025}. At each chunk, its diffusion transformer attends jointly to clean historical image latents and noisy new-view latents. The historical image latents anchor the appearance, and the model refreshes scene features across the active views to decode an updated set of Gaussians (\cref{fig:inference}). This lets ArgsWorld continue the same scene on a new camera orbit without generating bridging views.

Our \emph{3D Self-Forcing Distillation} trains the model in the same autoregressive setting used for inference (\cref{fig:distillation}). Completing all four denoising steps supplies the clean history for the next chunk. During training, we replay a randomly selected denoising step with gradients and apply distribution-matching distillation~\cite{yin2024dmd,yin2024dmd2} to the direct image predictions and to renderings of the updated Gaussian scene at both generation and additional cameras. The training mixture includes single-view images and posed multi-view sequences.

Our contributions are threefold:
\begin{itemize}
    \item \textbf{A unified 3DGS world model.} ArgsWorld uses one camera-conditioned autoregressive diffusion 3DGS model to reconstruct scenes from posed images, generate scenes from image or text prompts, and extrapolate scenes along an extended trajectory.
    \item \textbf{Consistent autoregressive scene expansion.} ArgsWorld grows a Gaussian world one chunk at a time, using completed generations as fixed visual context. Joint scene decoding and rendering consistency supervision encourage consistent renderings at historical viewpoints while allowing the scene representation to evolve as new regions are generated.
    \item \textbf{3D Self-Forcing Distillation.} Training on self-generated histories and matching distributions in both the image and rendering branches yields a four-step autoregressive sampler with explicit 3DGS output.
\end{itemize}

\section{Related Work}
\label{sec:RelatedWork}

\paragraph{Video generation and autoregressive world models.}
Video foundation models such as Wan~\cite{wan2025wan} supply visual priors for controllable world generation. Existing approaches condition generation on actions~\cite{bruce2024genie} or camera trajectories~\cite{he2025cameractrl,wang2024motionctrl,wang2026taming}. Training autoregressive video diffusion on ground-truth histories can lead to error accumulation on generated histories at inference. Self Forcing~\cite{huang2026self} addresses this mismatch by training on self-generated rollouts with sequence-level supervision. BiWM~\cite{rui2026biwm} retains bidirectional attention and combines camera-control adaptation with few-step distillation. CausVid and LongLive accelerate causal rollouts~\cite{yin2025causvid,yang2026longlive}, while WorldMem retrieves past views for consistency~\cite{xiao2025worldmem}. SolarWM~\cite{huang2026solarwm} scales long-horizon video training, and InSpatio-World~\cite{inspatio2026world} uses state-anchored autoregression for interactive 4D simulation. ArgsWorld uses self-generated histories to condition a bidirectional generator that jointly predicts images and updates an explicit Gaussian scene.

\paragraph{Gaussian scene reconstruction and generation.}
NeRF~\cite{mildenhall2020nerf} and 3D Gaussian Splatting~\cite{kerbl2023gaussians} enable differentiable rendering, with 3DGS providing explicit primitives. Feed-forward models estimate multiview geometry~\cite{wang2024dust3r,wang2025vggt} or synthesize novel views~\cite{wang2021ibrnet,jin2025lvsm}. pixelSplat, MVSplat, and DepthSplat reconstruct Gaussians from posed images~\cite{charatan2024pixelsplat,chen2024mvsplat,xu2025depthsplat}. AnySplat~\cite{jiang2025anysplat} handles uncalibrated image collections by predicting Gaussian primitives and camera parameters jointly. Diffusion priors support novel-view synthesis~\cite{liu2023zero123,gao2024cat3d,yu2025viewcrafter} and text-driven 3D generation~\cite{poole2023dreamfusion,hollein2023text2room}. FlashWorld~\cite{flashworld2025} integrates Gaussian decoding into video diffusion for image- or text-conditioned generation. For extrapolation, WonderWorld~\cite{yu2025wonderworld} constructs Gaussian surfels from outpainted views, and LucidDreamer~\cite{chung2025luciddreamer} uses diffusion-guided completion and alignment. WonderTurbo~\cite{ni2025wonderturbo} combines image inpainting and depth completion with incremental fusion of newly predicted 3D Gaussians. ArgsWorld revises the scene jointly across historical and new views within a single autoregressive generator.

\paragraph{Distribution-matching distillation.}
Distribution Matching Distillation (DMD)~\cite{yin2024dmd} trains a fast generator using the difference between the scores of a pretrained teacher and an online model of the generated distribution. DMD2~\cite{yin2024dmd2} improves score estimation through a two-time-scale update rule and adds adversarial supervision. FlashWorld~\cite{flashworld2025} uses distribution matching to transfer its multi-view image prior to Gaussian renderings. BiWM~\cite{rui2026biwm} distills bidirectional video generation from self-generated histories. ArgsWorld applies distribution matching to the direct image sequence and to renderings of the updated scene at two sets of cameras. This supplies training signals for both image synthesis and geometry through the same rollout.

\section{Method}
\label{sec:method}

ArgsWorld takes an ordered sequence of calibrated cameras, optional posed reference images, and an optional text prompt. It produces an explicit 3D Gaussian scene that can be rendered at additional query cameras. We initialize the model from a 3D adaptation of a video diffusion transformer (\cref{sec:adaptation}), then extend it to generate views in successive chunks (\cref{sec:ar_inference}). We train this autoregressive model by distilling its own rollouts in both image and rendering space (\cref{sec:self_forcing}).

\subsection{3D Adaptation}
\label{sec:adaptation}

The backbone is a camera-conditioned Wan2.2-TI2V-5B diffusion transformer (DiT)~\cite{wan2025wan}. We encode each RGB frame independently with the same frozen variational autoencoder (VAE), obtaining one normalized spatial latent grid per camera pose. Compression is spatial within each frame, so four input frames yield four latent grids. This preserves view-specific structure, such as contours and occlusion boundaries, and keeps it aligned with the corresponding camera rays for 3D decoding. For a camera ray with origin $\mathbf{o}$ and unit direction $\mathbf{d}$, the six-channel ray map concatenates $\mathbf{d}$ and $\mathbf{o}-(\mathbf{o}^{\top}\mathbf{d})\mathbf{d}$. The DiT takes noisy RGB latents, ray maps, and reference latents concatenated along the channel axis, with text supplied through cross-attention. Reference latents occupy the slots of their calibrated cameras, and unobserved slots are zero-filled.

\paragraph{Image and Gaussian outputs.}
We form the noisy latent $x_\sigma=(1-\sigma)z+\sigma\epsilon$ from a clean latent $z$ and Gaussian noise $\epsilon\sim\mathcal{N}(0,\mathbf{I})$. The DiT predicts a 48-channel velocity $v_\theta$ and a 1024-channel feature map per view. With the flow-matching velocity target $\epsilon-z$~\cite{lipman2023flowmatching}, the clean prediction is
\begin{equation}
    \hat z = x_\sigma - \sigma v_\theta .
    \label{eq:direct_rgb}
\end{equation}
A separate Gaussian decoder takes the predicted features, the detached clean latents, and the cameras. It uses a trainable copy of the Wan VAE decoder with feature injection and a pixel-aligned head that predicts two Gaussians per grid location. Each center is obtained from a subpixel ray offset and positive depth. The remaining outputs are opacity, anisotropic scale, rotation, and spherical harmonic appearance coefficients~\cite{kerbl2023gaussians}.

\paragraph{Initialization.}
We use the pretrained checkpoints from FlashWorld~\cite{flashworld2025}. The few-step checkpoint initializes the student DiT and Gaussian decoder, and the 3D-adapted diffusion checkpoint provides the frozen teacher.

\subsection{Chunk-wise Autoregressive Inference}
\label{sec:ar_inference}

We split the trajectory into ordered camera chunks $C_1,\ldots,C_B$. At chunk $b$, the state contains the cameras $C_{<b}$, the final direct RGB latents $H_{b-1}$, and the most recently computed features for those cameras. For the first chunk the history is empty, and any reference images enter through the reference channels at their matching camera slots. The model initializes the noisy RGB stream for the current chunk from Gaussian noise and the current feedback channels from zeros. This shared input format supports reconstruction with several posed references, generation with optional image or text prompts, and extrapolation by continuing a saved rollout with new cameras.

\paragraph{Bidirectional attention and scene update.}
At step $j$, we prepend $H_{b-1}$ to the noisy current latents $x_{b,j}$. Historical tokens receive timestep zero, and current tokens receive the noise level $\sigma_j$. The stored RGB latents and features also enter their conditioning channels in \cref{fig:inference}. With bidirectional attention over $C_{\leq b}$, the DiT predicts clean latents $\hat z_{b,j}$ for the current views via \cref{eq:direct_rgb} and refreshes features $F_{b,j}$ for all views. The decoder $\mathcal{D}_{\mathrm{GS}}$ rebuilds the scene using the features and the combined sequence $Z_{b,j}=[H_{b-1};\hat z_{b,j}]$,
\begin{equation}
    G_{b,j} = \mathcal{D}_{\mathrm{GS}}\!\left(
        F_{b,j}, \operatorname{sg}(Z_{b,j}), C_{\leq b}\right),
    \label{eq:gaussian_update}
\end{equation}
where $\operatorname{sg}$ denotes stop-gradient. For ray construction and decoding, we center the camera prefix at the first camera and normalize by its spatial extent. We then map Gaussian centers and scales back to the fixed scene frame. The clean RGB history anchors appearance during each update. Newly generated views can refine geometry in overlapping regions through the refreshed historical features. The decoder uses the current-view features to add geometry for regions revealed by the new cameras.

\paragraph{Rendering feedback and history.}
We write $\bar z(G,C)=\mathcal{E}(\mathcal{R}(G,C))$ for the latent obtained by rendering a scene $G$ at cameras $C$ and encoding the resulting views with the frozen VAE encoder $\mathcal{E}$. Between denoising steps, the current-view render latents $\bar z(G_{b,j},C_b)$ and features feed back into the DiT. We re-noise those render latents for the next step,
\begin{equation}
    x_{b,j+1} = (1-\sigma_{j+1})\,\bar z(G_{b,j},C_b)
                 + \sigma_{j+1}\epsilon ,
    \label{eq:render_feedback}
\end{equation}
with fresh Gaussian noise $\epsilon$. After $S=4$ steps, we carry $H_b=Z_{b,S}$ and $F_{b,S}$ into the next chunk and use $G_{b,S}$ as the scene for rendering.

\subsection{3D Self-Forcing Distillation}
\label{sec:self_forcing}

\begin{figure}[t]
    \centering
    \includegraphics[width=\columnwidth]{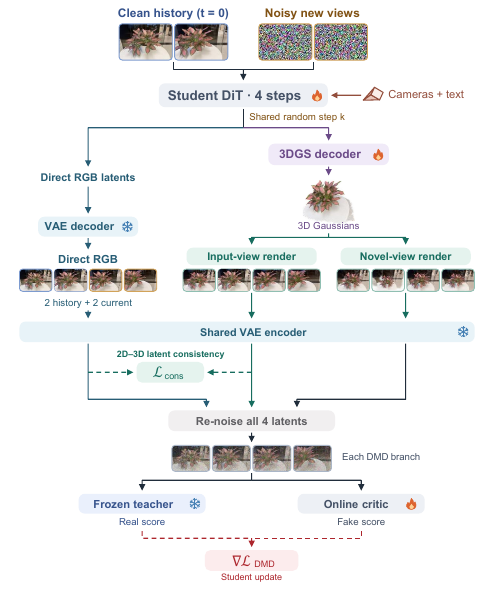}
    \caption{\textbf{3D Self-Forcing Distillation.}
    We replay a shared denoising step in each chunk and apply DMD to the direct RGB latents and both rendering branches. The input-view render latents also match the final chunk's clean RGB context.}
    \label{fig:distillation}
\end{figure}

\paragraph{Training on generated histories.}
Following Self Forcing~\cite{huang2026self} and BiWM~\cite{rui2026biwm}, we train the student on its own rollouts. Each chunk completes all $S$ denoising steps without gradients to supply the history for the next chunk. We sample a shared step $k$ uniformly from $\{1,\ldots,S\}$ and replay it with gradients from cached inputs. Appendix \Cref{sec:training_algorithm} gives the training pseudocode.

\paragraph{Three distillation branches.}
We apply DMD to the direct predictions from all chunks and to two sets of views rendered from the final chunk's scene (\cref{fig:distillation}).
\begin{equation}
    \begin{alignedat}{2}
        z_{\mathrm{rgb}} &= [\hat z_{1,k};\ldots;\hat z_{B,k}],
            &\quad& \mbox{direct RGB} \\
        z_{\mathrm{in}}  &= \bar z(G_{B,k},C_{\leq B}),
            && \mbox{input-view renders} \\
        z_{\mathrm{nov}} &= \bar z(G_{B,k},C_{\mathrm{nov}}),
            && \mbox{novel-view renders}
    \end{alignedat}
    \label{eq:dmd_branches}
\end{equation}
Here $C_{\leq B}$ contains all cameras in the rollout, and $C_{\mathrm{nov}}$ contains extra cameras sampled along the training trajectory. At novel views, both score networks condition only on cameras and text. The novel-view branch provides a training signal at cameras that contribute no image features to the Gaussian decoder, complementing supervision at the rollout cameras. Both rendering branches supervise the Gaussian decoder through differentiable rendering and VAE encoding.

\paragraph{Distribution matching.}
The frozen teacher and the online critic denoise the same detached student sample after re-noising at a random noise level. Let $\hat z_{\mathrm{real}}$ and $\hat z_{\mathrm{fake}}$ denote their clean estimates from \cref{eq:direct_rgb}. Following DMD~\cite{yin2024dmd,yin2024dmd2,flashworld2025}, the score-difference update in this clean-latent parameterization gives the gradient at an unobserved view $i$,
\begin{equation}
    \bigl[\nabla_z\mathcal{L}_{\mathrm{DMD}}\bigr]_i
        = -\frac{2}{Nw}
        \left(\hat z_{\mathrm{real},i}-\hat z_{\mathrm{fake},i}\right).
    \label{eq:dmd}
\end{equation}
Here $N$ is the number of latent elements in the branch output $z$, and $w$ is the mean absolute difference between $z$ and the teacher estimate, floored at $\delta$. At reference views, the encoded observation replaces the teacher estimate in $w$ and serves as the MSE target under the same reduction. The teacher uses text classifier-free guidance, and the branch-conditioned critic is trained by flow matching on detached student outputs.

\paragraph{2D--3D latent consistency.}
To preserve appearance as the scene is revised, we match the input-view render latents to the completed history and the final chunk's direct prediction used to decode the scene,
\begin{equation}
    \mathcal{L}_{\mathrm{cons}} =
        \operatorname{MSE}\!\left(z_{\mathrm{in}},[H_{B-1};\hat z_{B,k}]\right).
    \label{eq:latent_consistency}
\end{equation}


\begin{figure*}[t]
    \centering
    \includegraphics[width=\textwidth]{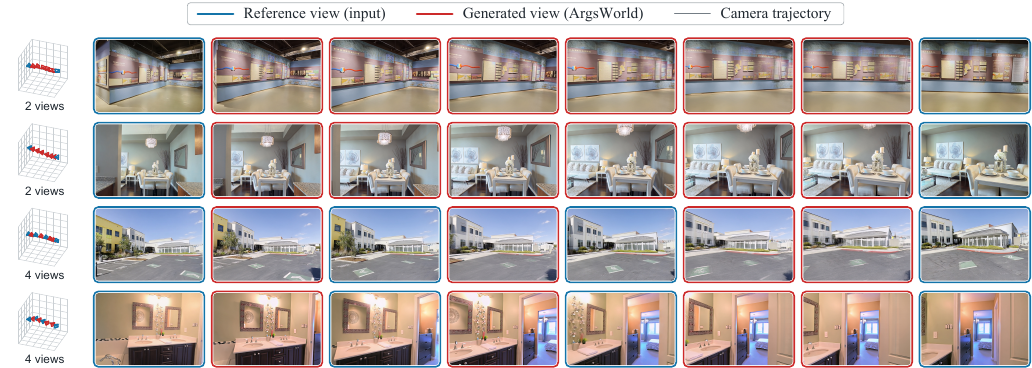}
    \caption{\textbf{Qualitative multi-view reconstruction.} ArgsWorld reconstructs scenes from two or four posed reference images. Blue borders mark inputs and red borders mark synthesized views. The diagrams at left show the camera trajectories.}
    \label{fig:reconstruction}
\end{figure*}

\section{Experiments}
\label{sec:experiments}

\paragraph{Training details.}
We initialize Wan2.2-TI2V-5B with FlashWorld weights~\cite{flashworld2025}. We alternate mixed-data and multi-view-only training stages. We use a curated, high-quality subset of DL3DV-10K~\cite{ling2024dl3dv}, RealEstate10K~\cite{zhou2018stereo}, ACID~\cite{liu2021infinitenature}, and SpatialVID~\cite{wang2026spatialvid} with sharp, high-fidelity clips, substantial camera motion, and high aesthetic quality~\cite{gao2026pixworld}. For single-view examples, we pair Echo-4o-Image~\cite{ye2025echo4o} with camera trajectories sampled randomly from WildRGB-D~\cite{xia2024wildrgbd}. In the final stage, we draw posed sequences and single-view images at a 2:1 ratio. To train for both generation and reconstruction, the first chunk of a multi-view rollout receives one to eight posed reference images from the same sequence.

We train on eight GPUs at $480{\times}704$, sampling two to five chunks of eight views per rollout. We weight the three DMD branches equally for both data sources. We evaluate the checkpoint at 4,200 global steps. Appendix \Cref{sec:implementation} gives the full training configuration. Appendix \Cref{sec:qualitative} shows additional qualitative results for reconstruction, generation, and extrapolation.

\subsection{Multi-view Reconstruction}
\label{sec:exp_reconstruction}

DL3DV-200 comprises 200 scenes from DL3DV-10K~\cite{ling2024dl3dv}. For each scene, we use 25 consecutive posed frames, provide four or eight as references, and render the remaining views at their ground-truth poses. We compare against ZipSplat, MVSplat, AnySplat, QuerySplat, and PixWorld. Inference runs at native resolution. We center-crop each prediction and ground-truth image to a square and resize both to $256{\times}256$ before computing PSNR, SSIM~\cite{wang2004ssim}, and LPIPS~\cite{zhang2018perceptual}. We average scores over target views and then over scenes.

\begin{table}[t]
    \centering
    \captionsetup{position=top}
    \caption{\textbf{Multi-view reconstruction on DL3DV-200.} Cell colors mark the \colorbox{TableBest}{best}, \colorbox{TableSecond}{second-best}, and \colorbox{TableThird}{third-best} scores in each metric column throughout the experiments.}
    \label{tab:reconstruction}
    \footnotesize
    \setlength{\tabcolsep}{1.7pt}
    \begin{tabular}{l*{6}{c}}
    \toprule
    & \multicolumn{3}{c}{4 views} & \multicolumn{3}{c}{8 views} \\
    \cmidrule(lr){2-4}\cmidrule(l){5-7}
    Method & PSNR$\uparrow$ & SSIM$\uparrow$ & LPIPS$\downarrow$
    & PSNR$\uparrow$ & SSIM$\uparrow$ & LPIPS$\downarrow$ \\
    \midrule
    ZipSplat~\cite{zipsplat2026} & 15.58 & 0.372 & 0.247 & 16.51 & 0.404 & 0.193 \\
    MVSplat~\cite{chen2024mvsplat} & \secondscore{20.25} & \bestscore{0.659} & 0.167 & 19.18 & \thirdscore{0.640} & 0.187 \\
    AnySplat~\cite{jiang2025anysplat} & 15.10 & 0.350 & 0.218 & 15.32 & 0.362 & 0.213 \\
    QuerySplat~\cite{querysplat2025} & 18.61 & 0.503 & \thirdscore{0.163} & \thirdscore{19.27} & 0.534 & \thirdscore{0.159} \\
    PixWorld~\cite{gao2026pixworld} & \thirdscore{19.85} & \thirdscore{0.610} & \secondscore{0.151} & \bestscore{22.53} & \bestscore{0.708} & \secondscore{0.121} \\
    \midrule
    \textbf{ArgsWorld} & \bestscore{20.68} & \secondscore{0.657} & \bestscore{0.121} & \secondscore{21.31} & \secondscore{0.694} & \bestscore{0.117} \\
    \bottomrule
    \end{tabular}
\end{table}

ArgsWorld achieves the lowest LPIPS with both four and eight references ($0.121$ and $0.117$). With four references, it also attains the highest PSNR ($20.68$\,dB) and the second-highest SSIM ($0.657$) (\cref{tab:reconstruction}). \Cref{fig:reconstruction} shows examples with two and four reference views.

\subsection{Single-view Scene Generation}
\label{sec:exp_generation}

\paragraph{Experimental setup.}
For each DL3DV-200 scene, we provide one reference image and a trajectory of eight camera poses. ArgsWorld generates eight views in one chunk, and all baselines generate the 8 requested views in one rollout except Gen3C and Gen3R, which generate 9 views in one rollout. We compare Gaussian renderings from ArgsWorld, FlashWorld, and PixWorld and RGB predictions from Geometry Forcing, Gen3C, and Gen3R with the corresponding ground-truth frames using PSNR, SSIM, and LPIPS. We evaluate inference speed on a single NVIDIA RTX 6000D, including Gaussian rendering where applicable, while excluding model loading and output writes.

\begin{figure*}[t]
    \centering
    \includegraphics[width=\textwidth]{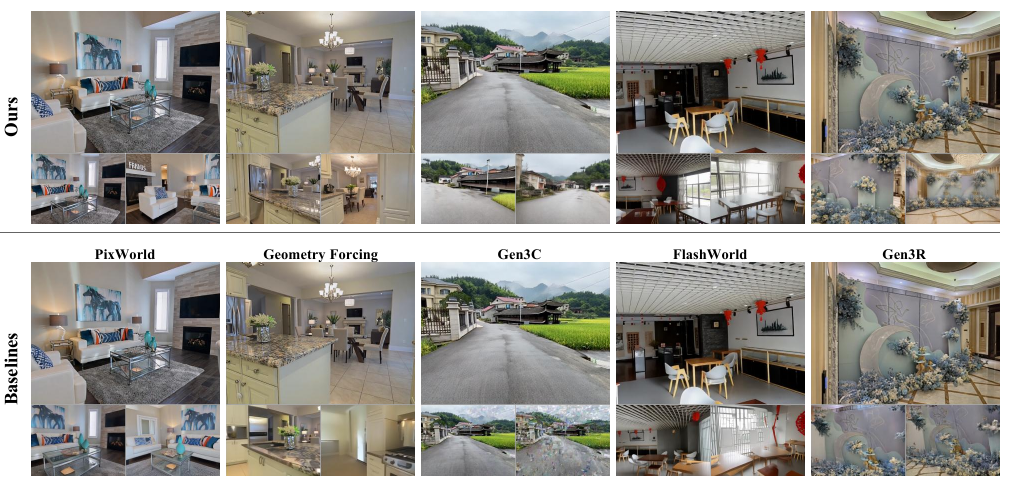}
    \caption{\textbf{Qualitative single-view scene generation.} Each column compares ArgsWorld with the named baseline using the same reference image. The smaller panels show views 8 and 16 from direct 16-view generation.}
    \label{fig:generation}
\end{figure*}

\begin{table}[t]
    \centering
    \captionsetup{position=top}
    \caption{\textbf{Quantitative comparison of single-image 3D scene generation on DL3DV-200.}}
    \label{tab:generation}
    \footnotesize
    \setlength{\tabcolsep}{2.7pt}
    \begin{tabular}{l*{4}{c}}
    \toprule
    Method & PSNR$\uparrow$ & SSIM$\uparrow$ & LPIPS$\downarrow$ & Time (s)$\downarrow$ \\
    \midrule
    Geometry Forcing~\cite{wu2026geometryforcing} & 8.4218 & 0.1363 & \thirdscore{0.6206} & 19.28 \\
    Gen3C~\cite{ren2025gen3c} & 10.5259 & 0.2968 & 0.7110 & 65.22 \\
    Gen3R~\cite{huang2026gen3r} & 10.5046 & \thirdscore{0.3036} & 0.6561 & 26.21 \\
    FlashWorld~\cite{flashworld2025} & \thirdscore{11.3820} & \secondscore{0.3041} & \bestscore{0.5834} & \thirdscore{2.55} \\
    PixWorld~\cite{gao2026pixworld} & \secondscore{11.4295} & 0.2957 & 0.6504 & \bestscore{1.37} \\
    \midrule
    \textbf{ArgsWorld} & \bestscore{11.6104} & \bestscore{0.3054} & \secondscore{0.5939} & \secondscore{2.28} \\
    \bottomrule
    \end{tabular}
\end{table}

ArgsWorld achieves the highest PSNR ($11.6104$\,dB) and SSIM ($0.3054$), and the second-lowest LPIPS ($0.5939$) in \cref{tab:generation}. \Cref{fig:generation} shows qualitative comparisons on separate 16-view runs.

\begin{figure*}[t]
    \centering
    \includegraphics[width=\textwidth]{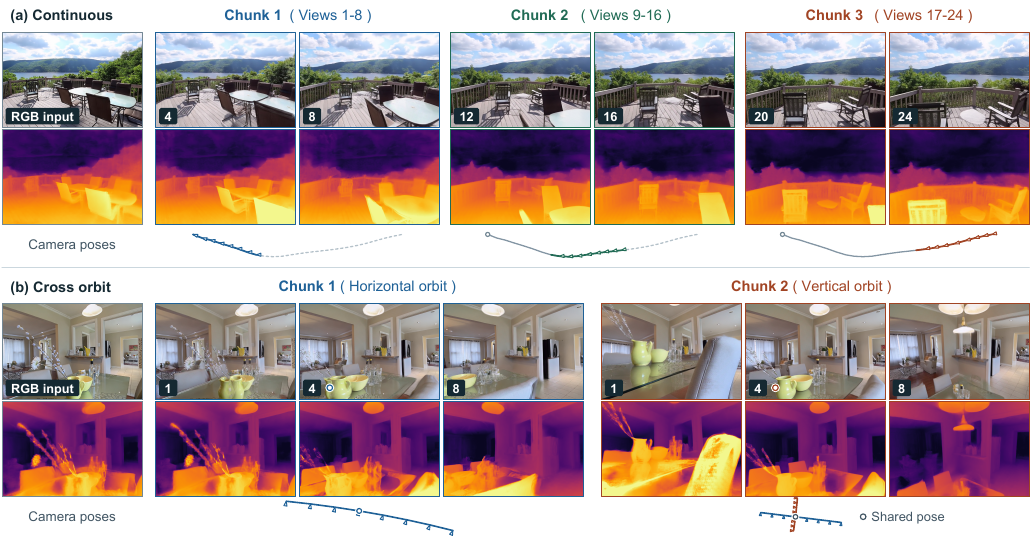}
    \caption{\textbf{Autoregressive extrapolation from one image.} We show RGB and depth for selected views. (a) Three eight-view chunks follow a continuous trajectory. (b) A second chunk switches from a horizontal to a vertical orbit. The first orbit's endpoint and the second orbit's start are separated, with a shared interior camera pose marked on both paths.}
    \label{fig:extrapolation}
\end{figure*}

\subsection{Autoregressive Scene Extrapolation}
\label{sec:exp_extrapolation}

\paragraph{Experimental setup.}
For each DL3DV-200 scene, we provide one reference image and a trajectory of 16 camera poses. ArgsWorld splits the trajectory into two consecutive eight-view chunks. It uses the generated history from the first chunk to synthesize the second and update the Gaussian scene. All baselines generate the 16 requested views in one rollout except Gen3C and Gen3R, which generate 17 views in one rollout. We score Gaussian renderings from ArgsWorld, FlashWorld, and PixWorld and RGB predictions from Geometry Forcing, Gen3C, and Gen3R against the corresponding ground-truth frames using PSNR, SSIM, and LPIPS.

\begin{table}[t]
    \centering
    \captionsetup{position=top}
    \caption{\textbf{Quantitative comparison of single-image autoregressive 3D scene extrapolation on DL3DV-200.}}
    \label{tab:extrapolation}
    \footnotesize
    \setlength{\tabcolsep}{2.7pt}
    \begin{tabular}{l*{4}{c}}
    \toprule
    Method & PSNR$\uparrow$ & SSIM$\uparrow$ & LPIPS$\downarrow$ & Time (s)$\downarrow$ \\
    \midrule
    Geometry Forcing~\cite{wu2026geometryforcing} & 8.9931 & 0.1703 & \thirdscore{0.6042} & 57.47 \\
    Gen3C~\cite{ren2025gen3c} & 10.5526 & \thirdscore{0.3132} & 0.7042 & 107.54 \\
    Gen3R~\cite{huang2026gen3r} & 10.8525 & \secondscore{0.3137} & 0.6381 & 56.12 \\
    FlashWorld~\cite{flashworld2025} & \secondscore{11.7511} & 0.3077 & \bestscore{0.5620} & \secondscore{4.96} \\
    PixWorld~\cite{gao2026pixworld} & \thirdscore{11.7496} & 0.3095 & 0.6428 & \bestscore{3.04} \\
    \midrule
    \textbf{ArgsWorld} & \bestscore{11.8344} & \bestscore{0.3140} & \secondscore{0.5910} & \thirdscore{6.91} \\
    \bottomrule
    \end{tabular}
\end{table}

\begin{table}[t]
    \centering
    \captionsetup{position=top}
    \caption{\textbf{Text-conditioned generation on T3Bench-200.}}
    \label{tab:t3bench}
    \footnotesize
    \setlength{\tabcolsep}{2pt}
    \begin{tabular}{l*{5}{c}}
    \toprule
    & \multicolumn{2}{c}{\textit{Q-Align}} & \multicolumn{3}{c}{\textit{CLIP}} \\
    \cmidrule(lr){2-3}\cmidrule(l){4-6}
    Method & IQA$\uparrow$ & IAA$\uparrow$ & IQA+$\uparrow$ & Aesthetic$\uparrow$ & Score$\uparrow$ \\
    \midrule
    FlashWorld~\cite{flashworld2025} & \secondscore{4.42} & \thirdscore{2.46} & \secondscore{0.65} & \bestscore{4.94} & \secondscore{32.53} \\
    PixWorld~\cite{gao2026pixworld} & \thirdscore{2.91} & \secondscore{2.48} & \thirdscore{0.48} & \thirdscore{4.61} & \thirdscore{26.76} \\
    \midrule
    \textbf{ArgsWorld} (Generation) & \bestscore{4.44} & \bestscore{2.51} & \bestscore{0.67} & \secondscore{4.93} & \bestscore{33.10} \\
    \bottomrule
    \end{tabular}
\end{table}

\begin{table*}[!t]
    \centering
    \captionsetup{position=top}
    \caption{\textbf{Static world generation on WorldScore.} Average is the unweighted mean of the six metrics.}
    \label{tab:worldscore}
    \small
    \setlength{\tabcolsep}{5.6pt}
    \begin{tabular}{l*{6}{c}|c}
    \toprule
    Method & \shortstack{Camera\\Control} & \shortstack{Object\\Control}
    & \shortstack{Content\\Alignment} & \shortstack{3D\\Consistency}
    & \shortstack{Photometric\\Consistency} & \shortstack{Subjective\\Quality}
    & \textbf{Average} \\
    \midrule
    DynamiCrafter~\cite{xing2024dynamicrafter} & 25.15 & \thirdscore{47.36} & 25.00 & 72.90 & 60.95 & 54.40 & 47.63 \\
    VideoCrafter1-I2V~\cite{chen2023videocrafter1} & 25.46 & 24.25 & 35.27 & 74.42 & 73.89 & 54.85 & 48.02 \\
    Wan-2.1~\cite{wan2025wan} & 23.53 & 40.32 & 45.44 & 78.74 & 78.36 & \secondscore{59.38} & 54.30 \\
    WonderJourney~\cite{yu2024wonderjourney} & \secondscore{84.60} & 37.10 & 35.54 & 80.60 & 79.03 & \bestscore{66.56} & 63.91 \\
    FlashWorld~\cite{flashworld2025} & \thirdscore84.43 & \secondscore{50.28} & \secondscore{56.54} & \secondscore{85.87} & \secondscore{86.72} & 52.75 & \thirdscore{69.43} \\
    PixWorld~\cite{gao2026pixworld} & \bestscore{91.08} & 46.25 & \thirdscore{55.27} & \bestscore{91.39} & \bestscore{93.84} & 52.36 & \secondscore{71.70} \\
    \midrule
    \textbf{ArgsWorld} & \bestscore{91.08} & \bestscore{50.60} & \bestscore{65.86} & \thirdscore{84.70} & \thirdscore{82.40} & \thirdscore{57.83} & \bestscore{72.08} \\
    \bottomrule
    \end{tabular}
\end{table*}

ArgsWorld achieves the highest PSNR ($11.8344$\,dB) and SSIM ($0.3140$), and the second-lowest LPIPS ($0.5910$) in \cref{tab:extrapolation}.

\paragraph{Continuation across disconnected endpoints.}
\Cref{fig:extrapolation}(a) illustrates a longer rollout over three consecutive eight-view chunks. In the cross-orbit example in \cref{fig:extrapolation}(b), the next chunk switches from a horizontal orbit to a vertical orbit whose starting camera differs from the previous endpoint. The two orbits share their fourth pose. ArgsWorld carries its generated history into the vertical orbit and updates the same Gaussian scene without generating bridging views.

\subsection{Text-conditioned Generation on T3Bench-200}
\label{sec:exp_t3bench}

\paragraph{Experimental setup.}
T3Bench-200 contains 200 text prompts sampled from T3Bench~\cite{he2024t3bench}, covering single objects, objects with surroundings, and multi-object scenes. For each prompt, ArgsWorld generates 16 views in one chunk. FlashWorld and PixWorld also generate 16 views in a single rollout. We assess image quality and aesthetics with Q-Align IQA and IAA~\cite{wu2024qalign}, perceptual quality with CLIP-IQA+~\cite{wang2023clipiqa}, and aesthetic appeal with the CLIP aesthetic predictor. We use CLIPScore to measure alignment with the input prompt~\cite{hessel2021clipscore}.

\begin{figure}[!t]
    \centering
    \includegraphics[pagebox=cropbox,width=.97\columnwidth]{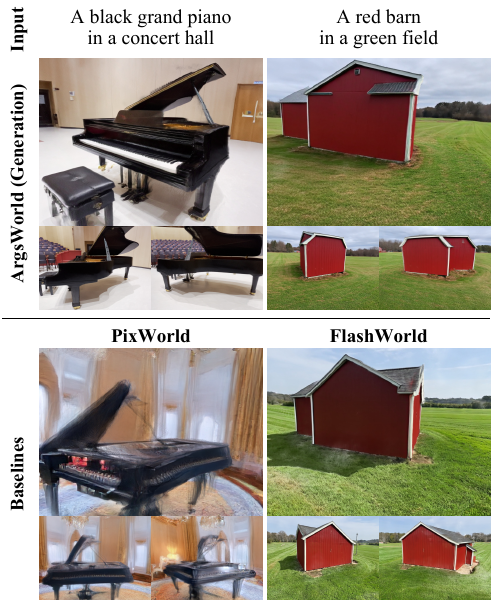}
    \caption{\textbf{Qualitative text-conditioned generation.} The piano example compares ArgsWorld with PixWorld, and the barn example compares it with FlashWorld.}
    \label{fig:t3bench}
\end{figure}

ArgsWorld ranks first on four of five metrics in \cref{tab:t3bench}, including prompt alignment with a CLIPScore of $33.10$. \Cref{fig:t3bench} shows qualitative comparisons for two prompts.

\subsection{Evaluation on WorldScore Benchmark}
\label{sec:exp_worldscore}

\paragraph{Experimental setup.}
We evaluate ArgsWorld on the 2,000 cases in the static split of WorldScore~\cite{duan2025worldscore}. Each case provides a reference image, scene descriptions, and a target camera trajectory. We use the published baseline scores~\cite{duan2025worldscore,gao2026pixworld}. Camera Control measures trajectory adherence, Object Control checks for requested objects, and Content Alignment measures agreement with the prompt. We also report 3D Consistency for cross-view geometry, Photometric Consistency for cross-view appearance, and Subjective Quality for perceptual quality. We compute the reported Average as the unweighted mean of these six metrics.

ArgsWorld achieves the highest six-metric average ($72.08$) in \cref{tab:worldscore}. It shares the highest camera control score ($91.08$) with PixWorld and ranks first in object control ($50.60$) and content alignment ($65.86$).

\subsection{Ablation Study}
\label{sec:exp_ablation}

We evaluate both variants with the single-view DL3DV-200 protocols in \cref{sec:exp_generation,sec:exp_extrapolation}. Starting from one reference image, the model generates one eight-view chunk for generation or two successive eight-view chunks for extrapolation. We score Gaussian renderings at the target cameras.

\begin{table}[t]
    \centering
    \captionsetup{position=top}
    \caption{\textbf{Ablations on DL3DV-200.} Both settings start from one reference image.}
    \label{tab:ablation}
    \footnotesize
    \setlength{\tabcolsep}{4pt}
    \begin{tabular}{lccc}
    \toprule
    Method & PSNR$\uparrow$ & SSIM$\uparrow$ & LPIPS$\downarrow$ \\
    \midrule
    \multicolumn{4}{l}{\textit{Generation (8 views, one chunk)}} \\
    \textbf{ArgsWorld} & \bestscore{11.6104} & \secondscore{0.3054} & \bestscore{0.5939} \\
    Direct RGB DMD only & \secondscore{10.8163} & \bestscore{0.3152} & \secondscore{0.6383} \\
    Causal attention & \thirdscore{8.0975} & \thirdscore{0.1377} & \thirdscore{0.6538} \\
    \midrule
    \multicolumn{4}{l}{\textit{Extrapolation (16 views, two chunks)}} \\
    \textbf{ArgsWorld} & \bestscore{11.8344} & \bestscore{0.3140} & \bestscore{0.5910} \\
    Direct RGB DMD only & \secondscore{10.6788} & \secondscore{0.3121} & \secondscore{0.6484} \\
    Causal attention & \thirdscore{8.8900} & \thirdscore{0.1773} & \thirdscore{0.6485} \\
    \bottomrule
    \end{tabular}
\end{table}

\paragraph{Direct RGB DMD only.}
This variant applies DMD only to the direct RGB latents $z_{\mathrm{rgb}}$ in \cref{eq:dmd_branches}. In the full model, the $z_{\mathrm{in}}$ and $z_{\mathrm{nov}}$ branches supervise the Gaussian decoder through input-view and novel-view renderings. In \cref{tab:ablation}, adding them improves PSNR by $0.79$\,dB in generation and $1.16$\,dB in extrapolation, and reduces LPIPS by $0.0444$ and $0.0574$, respectively. The full model improves all three metrics in extrapolation.

\paragraph{Causal attention.}
The causal variant lets current views attend to earlier views and freezes the historical features used for Gaussian decoding. In the full model, bidirectional attention refreshes the historical features in $F_{b,j}$ used for scene decoding in \cref{eq:gaussian_update} and keeps the clean RGB history fixed. With causal attention, PSNR falls by $3.51$\,dB in generation and $2.94$\,dB in extrapolation, with lower SSIM in both settings. In extrapolation, the comparison reflects the effect of causal masking and frozen first-chunk features.

\section{Conclusion}
\label{sec:conclusion}

We present ArgsWorld, a unified autoregressive diffusion transformer for explicit 3D Gaussian scene modeling. Our framework integrates multi-view reconstruction, image- and text-conditioned scene generation, and camera-controlled extrapolation within a single generative architecture. Bidirectional attention refreshes scene features while retaining clean RGB history, and 3D Self-Forcing Distillation supervises RGB predictions and Gaussian renderings on self-generated rollouts. This strategy improves visual fidelity and cross-view consistency while enabling efficient four-step inference. Extensive experiments demonstrate state-of-the-art performance across multiple 3D scene modeling benchmarks, validating the effectiveness of our unified approach. These results establish autoregressive diffusion as a promising paradigm for constructing, generating, and continuously expanding explicit 3D worlds.

\FloatBarrier
{
    \small
    \bibliographystyle{ieeenat_fullname}
    \bibliography{main}
}

\appendix
\twocolumn[{%
    \centering
    {\Large\bfseries \thetitle\par}
    \vspace{0.6em}
    {\Large Appendix\par}
    \vspace{1.0em}
}]
\section{IMPLEMENTATION DETAILS}
\label{sec:implementation}

\paragraph{Architecture.}
The Wan2.2-TI2V-5B backbone has 30 transformer blocks and 24 attention heads of dimension 128. We remove temporal padding from the frozen VAE and sample a normalized 48-channel posterior latent independently for each view. The encoder downsamples by a factor of 16 along each image axis. At $480{\times}704$, the $30{\times}44$ latent grid becomes 330 tokens per view after $2{\times}2$ patchification. The student concatenates 48 noisy latent channels, six ray channels, 48 reference channels, 1024 feedback feature channels, and 48 feedback latent channels, for 1174 input channels. The teacher sees the first three groups, and the critic adds one channel carrying the branch identifier. We encode text with the frozen UMT5 encoder at a maximum length of 512 tokens.

\paragraph{Gaussian decoding and rendering.}
The reconstruction decoder adds a convolutional projection of the DiT features to the first decoder activation of the unnormalized clean latent. Its prediction head produces two Gaussians per spatial output location. A truncated exponential gives positive depth, and a sigmoid bounds opacity and the pixel-size-scaled anisotropic scales. Each Gaussian uses a unit quaternion and degree-two spherical harmonics. We rasterize RGB and depth in FP32. Multiscale depth regularization uses average pooling with kernel sizes $4$, $2$, and $1$, weighted by $4/7$, $2/7$, and $1/7$, respectively.

\paragraph{Data and conditioning.}
We sample posed sequences and single-view images at a $2{:}1$ ratio. Each rollout uses a uniformly sampled length of two to five eight-view chunks. Within the multi-view stream, the sampling weights for DL3DV-10K, RealEstate10K, ACID, and SpatialVID are $4{:}2{:}1{:}1$. The nominal input frame strides are $3$--$4$ for DL3DV-10K and ACID, $5$--$12$ for RealEstate10K, $3$--$5$ for SpatialVID, and $5$--$7$ for WildRGB-D. Each posed clip shares a center crop with a crop-scale factor drawn uniformly from $[0.8,1.0]$. We update the intrinsics for the crop and resize. For the novel-view branch, we sample a second trajectory of equal length at a smaller frame stride within the span of the rollout. Its camera slots carry no reference latents. Image conditioning is enabled with probability $0.7$, with prompt dropout of $0.1$ on image-conditioned examples. We sample the number of posed references uniformly from one to eight and select distinct slots uniformly in the first chunk. A single-view example uses the first slot for its reference image when image conditioning is enabled.

\paragraph{Distillation and optimization.}
The online critic starts from the same adapted FlashWorld checkpoint as the frozen teacher. The teacher uses text classifier-free guidance with scale $3$. We sample one of the four denoising steps uniformly for replay and share it across chunks. We average the three DMD losses and set the normalizer floor in \cref{eq:dmd} to $10^{-6}$. We sample DMD noise from scheduler entries within the nominal range $\sigma\in[0.02,0.98]$, with bounds tied to the replayed denoising step. The adversarial head projects the 1024-channel critic features to 128 channels and applies three stride-two residual blocks of widths 128, 256, and 512 to obtain per-view patch logits. For multi-view samples, we regularize these logits with an MSE between real predictions before and after adding Gaussian noise of standard deviation $0.05$ to the noisy real latents. The penalty has weight $10$ in the critic objective. The logistic adversarial terms for the student and critic each have weight $0.005$ per branch. The depth and opacity penalties each have weight $0.01$. The latent consistency term has weight $0.1$ through step 2,200, $0.4$ through step 3,000, $0.6$ through step 3,350, $1.0$ through step 4,000, and $0.4$ for the final 200 steps.

We use AdamW with peak learning rates of $10^{-6}$ for the student and Gaussian decoder and $5{\times}10^{-7}$ for the critic and adversarial head, with $(\beta_1,\beta_2)=(0.9,0.95)$, weight decay $10^{-6}$, and gradient clipping at $1$. Learning rates warm up over 1,000 global steps. After 100 critic-only steps, we update the student and decoder once every four global steps and the critic at every step. The global batch contains eight examples, one per GPU, with no gradient accumulation. We use BF16 autocasting, fully shard the DiTs, and checkpoint activations in the trainable networks and rendering pipeline.

\paragraph{Inference.}
Each chunk uses noise levels $(1,0.9,0.75,0.5)$, corresponding to zero-based indices $(0,250,500,750)$ in the flow-matching scheduler with shift $3$. The first camera and the spatial extent of the first chunk define the fixed session frame. At each denoising step, the local normalization divides camera offsets from the first camera by the largest offset norm in the prefix plus $10^{-2}$. We map centers and scales back to the session frame after decoding, and render RGB against white for feedback to the next denoising step. The final clean direct latents and refreshed features form the history for the next chunk.

\twocolumn[{\section{TRAINING ALGORITHM}
\label{sec:training_algorithm}

\begingroup
\newcommand{\trainingline}[2][0]{%
    \item \hspace*{#1em}\parbox[t]{\dimexpr\linewidth-#1em\relax}{\raggedright #2}%
}
\hrule
\vspace{4pt}
\noindent\textbf{3D Self-Forcing Distillation}\par
\vspace{4pt}
\hrule
\begin{enumerate}[label=\scriptsize\arabic*.,leftmargin=2em,labelsep=0.65em,topsep=5pt,itemsep=2pt]
    \trainingline{\textbf{Initialize} the student and Gaussian decoder from the few-step FlashWorld checkpoint.}
    \trainingline{Initialize the frozen teacher and trainable critic from the 3D-adapted diffusion checkpoint.}
    \trainingline{\textbf{for} global step $m=1,\ldots,M$ \textbf{do}}
    \trainingline[1]{Draw a training example following \cref{sec:implementation}. Sample $B\sim\mathrm{Unif}\{2,\ldots,5\}$ and a shared replay step $k\sim\mathrm{Unif}\{1,\ldots,4\}$.}
    \trainingline[1]{Prepare $B$ eight-view chunks, the denser trajectory $C_{\mathrm{nov}}$, text conditioning, and any first-chunk references.}
    \trainingline[1]{$u\gets (m>100)\wedge(m\bmod 4=0)$ \hfill \emph{Student update flag}}
    \trainingline[1]{Initialize the RGB and feature histories to empty.}
    \trainingline[1]{\textbf{for} chunk $b=1,\ldots,B$ \textbf{do}}
    \trainingline[2]{Initialize the chunk from Gaussian noise and zero feedback.}
    \trainingline[2]{\textbf{for} denoising step $j=1,\ldots,4$ \textbf{do} \hfill \emph{No gradients}}
    \trainingline[3]{Predict $\hat z_{b,j}$ and refresh $F_{b,j}$ using the clean prefix history and current inputs.}
    \trainingline[3]{Decode $G_{b,j}$ using $F_{b,j}$ and $\operatorname{sg}([H_{b-1};\hat z_{b,j}])$ (\cref{eq:gaussian_update}).}
    \trainingline[3]{At $j=k$, retain the outputs. If $u$ is true, also cache the step inputs for replay.}
    \trainingline[3]{If $j<4$, render and encode the current views, then re-noise them and pass the feedback to the next step (\cref{eq:render_feedback}).}
    \trainingline[2]{\textbf{end for}}
    \trainingline[2]{$H_b,F_b\gets \operatorname{sg}([H_{b-1};\hat z_{b,4}]),\operatorname{sg}(F_{b,4})$}
    \trainingline[2]{If $u$ is true, replay step $k$ from its cached inputs with gradients to obtain $\hat z_{b,k}$ and $G_{b,k}$. Otherwise use the outputs recorded at step $k$.}
    \trainingline[1]{\textbf{end for}}
    \trainingline[1]{$z_{\mathrm{rgb}}\gets[\hat z_{1,k};\ldots;\hat z_{B,k}]$}
    \trainingline[1]{$z_{\mathrm{in}},z_{\mathrm{nov}}\gets\bar z(G_{B,k},C_{\leq B}),\bar z(G_{B,k},C_{\mathrm{nov}})$}
    \trainingline[1]{\textbf{if} $u$ \textbf{then}}
    \trainingline[2]{Hold the critic and adversarial head fixed for this update. Sample a score noise level conditioned on $k$.}
    \trainingline[2]{For each branch, score the same re-noised latent with the teacher and branch-conditioned critic. Compute the DMD update using \cref{eq:dmd} and the reference-view anchors.}
    \trainingline[2]{Compute $\mathcal{L}_{\mathrm{cons}}=\operatorname{MSE}(z_{\mathrm{in}},[H_{B-1};\hat z_{B,k}])$.}
    \trainingline[2]{Update the student and Gaussian decoder using mean branch DMD, consistency, depth and opacity regularization, and the sum of logistic adversarial losses for multi-view examples.}
    \trainingline[1]{\textbf{end if}}
    \trainingline[1]{Sample a fresh score noise level conditioned on $k$ and re-noise the detached branch latents from the same rollout.}
    \trainingline[1]{Update the critic by flow matching to $\epsilon-\operatorname{sg}(z)$ in each branch. For multi-view examples, also update the critic and adversarial head with the per-branch logistic losses on camera-matched data and the real-logit perturbation penalty.}
    \trainingline{\textbf{end for}}
\end{enumerate}
\hrule
\endgroup
}]

\onecolumn
\ifdefined\nolinenumbers
    \nolinenumbers
\fi

\section{ADDITIONAL QUALITATIVE RESULTS}
\label{sec:qualitative}


\subsection{Multi-view Reconstruction}
\par\vspace*{\fill}%
\noindent\makebox[\textwidth][c]{%
    \includegraphics[pagebox=cropbox,trim=24bp 48bp 18bp 45bp,clip,width=\textwidth,
                     height=.9\textheight,keepaspectratio]{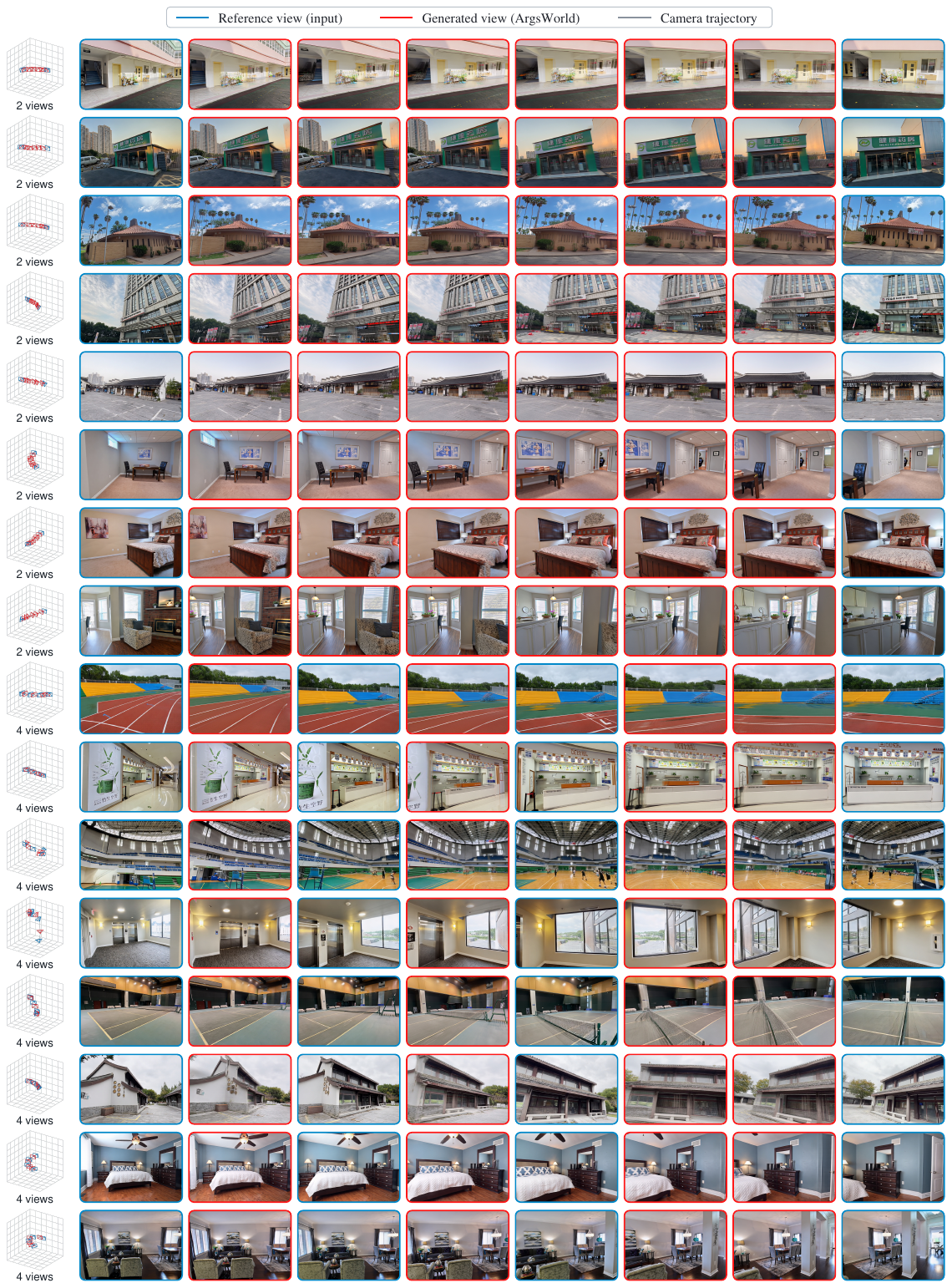}%
}\par\vfill

\newpage
\subsection{Text-conditioned Generation}
\par\vspace*{\fill}%
\noindent\makebox[\textwidth][c]{%
    \includegraphics[pagebox=cropbox,trim=0bp 0bp 0bp 0bp,clip,width=\textwidth,
                     height=.9\textheight,keepaspectratio]{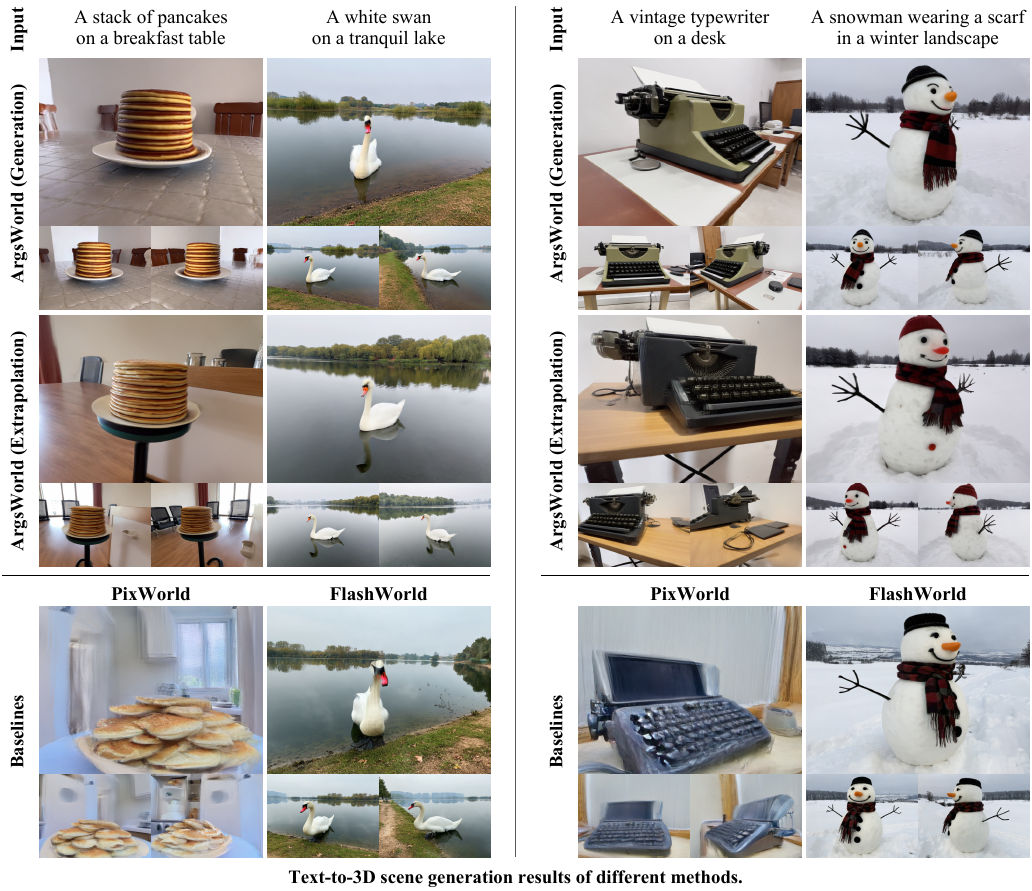}%
}\par\vfill

\newpage
\subsection{Autoregressive Extrapolation}
\subsubsection{Continuous Camera Trajectories}
\par\vspace*{\fill}%
\noindent\makebox[\textwidth][c]{%
    \includegraphics[pagebox=cropbox,trim=0bp 10bp 0bp 11bp,clip,width=\textwidth,
                     height=.9\textheight,keepaspectratio]{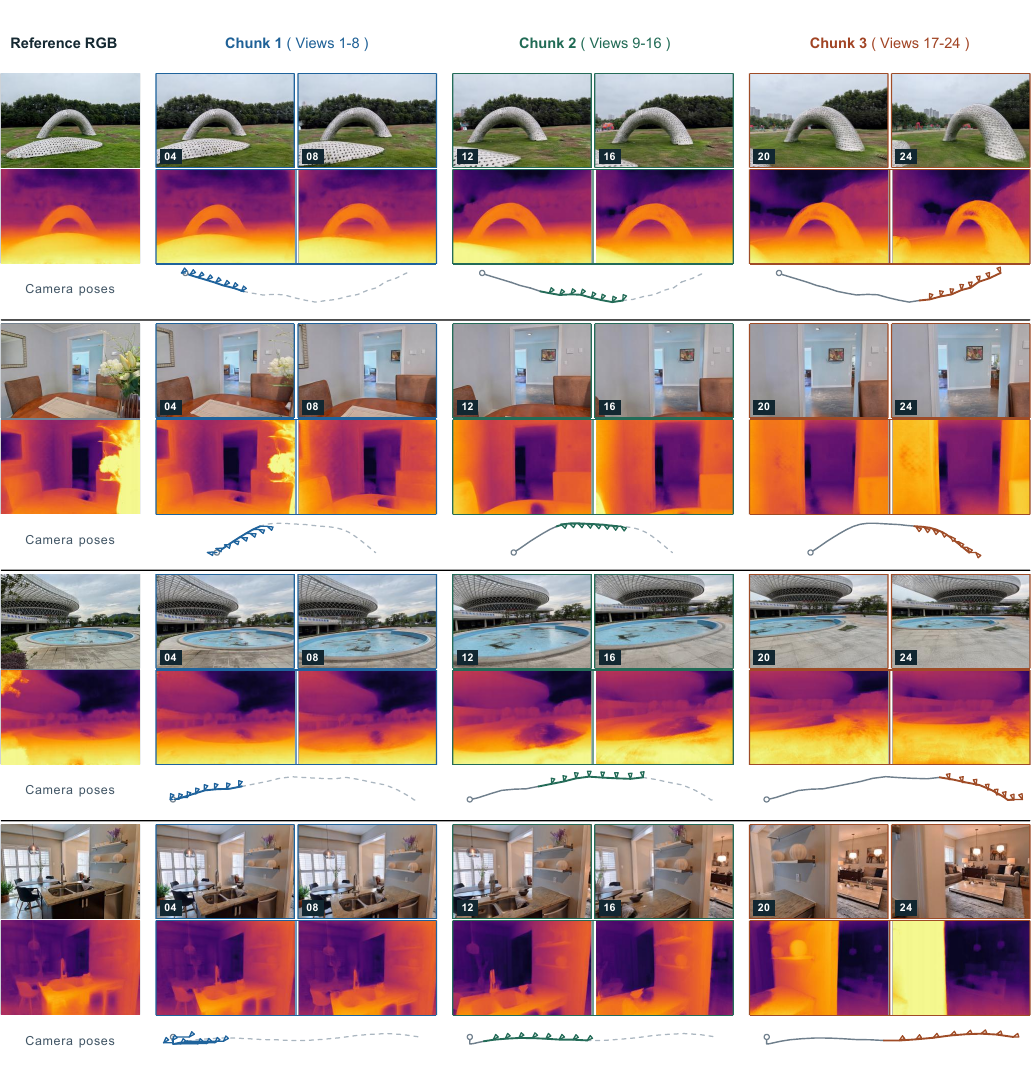}%
}\par\vfill

\newpage
\par\vspace*{\fill}%
\noindent\makebox[\textwidth][c]{%
    \includegraphics[pagebox=cropbox,trim=0bp 80bp 0bp 36bp,clip,width=\textwidth,
                     height=.9\textheight,keepaspectratio]{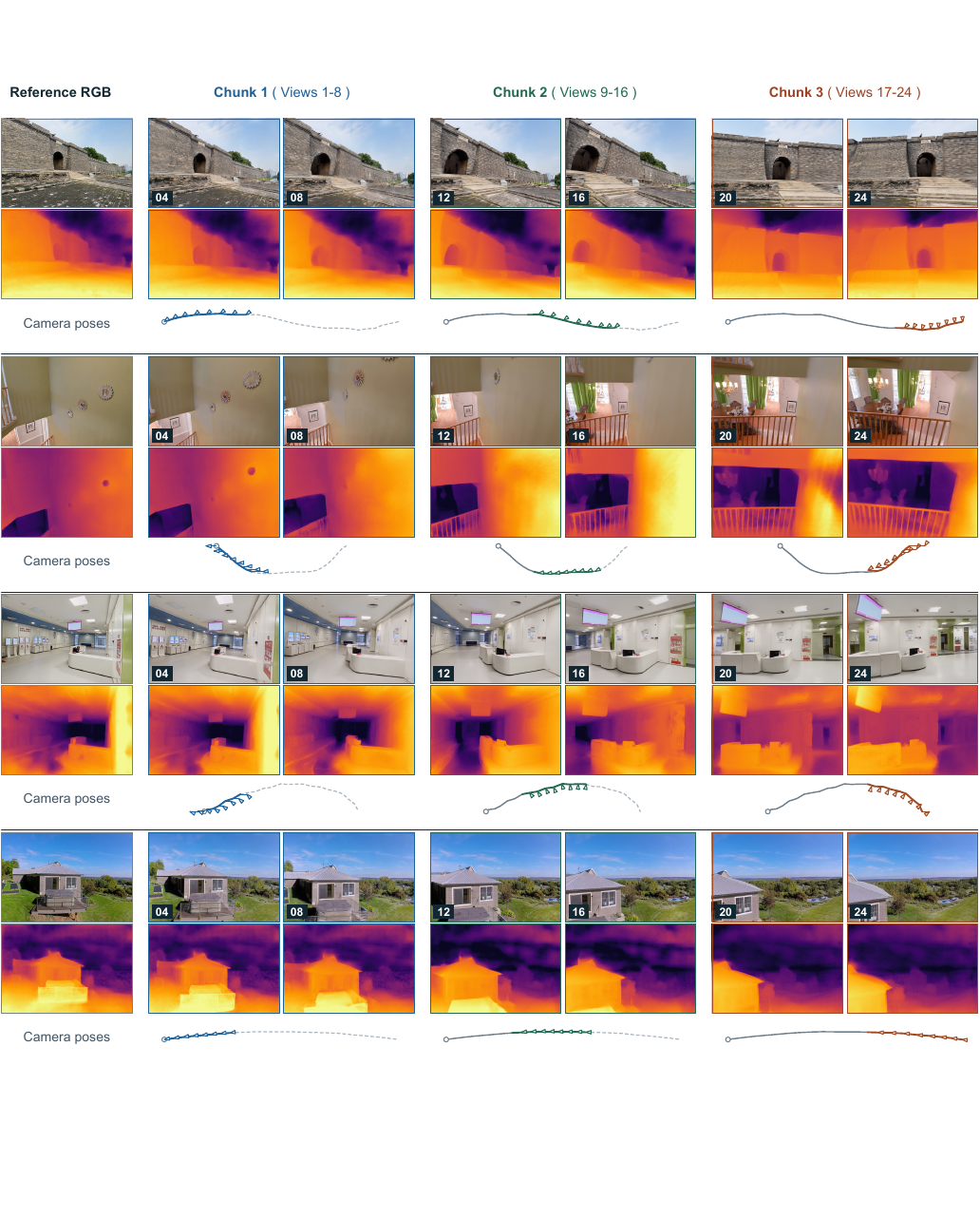}%
}\par\vfill

\newpage
\subsubsection{Horizontal-to-Vertical Camera Orbits}
\par\vspace*{\fill}%
\noindent\makebox[\textwidth][c]{%
    \includegraphics[pagebox=cropbox,trim=0bp 7bp 0bp 6bp,clip,width=\textwidth,
                     height=.9\textheight,keepaspectratio]{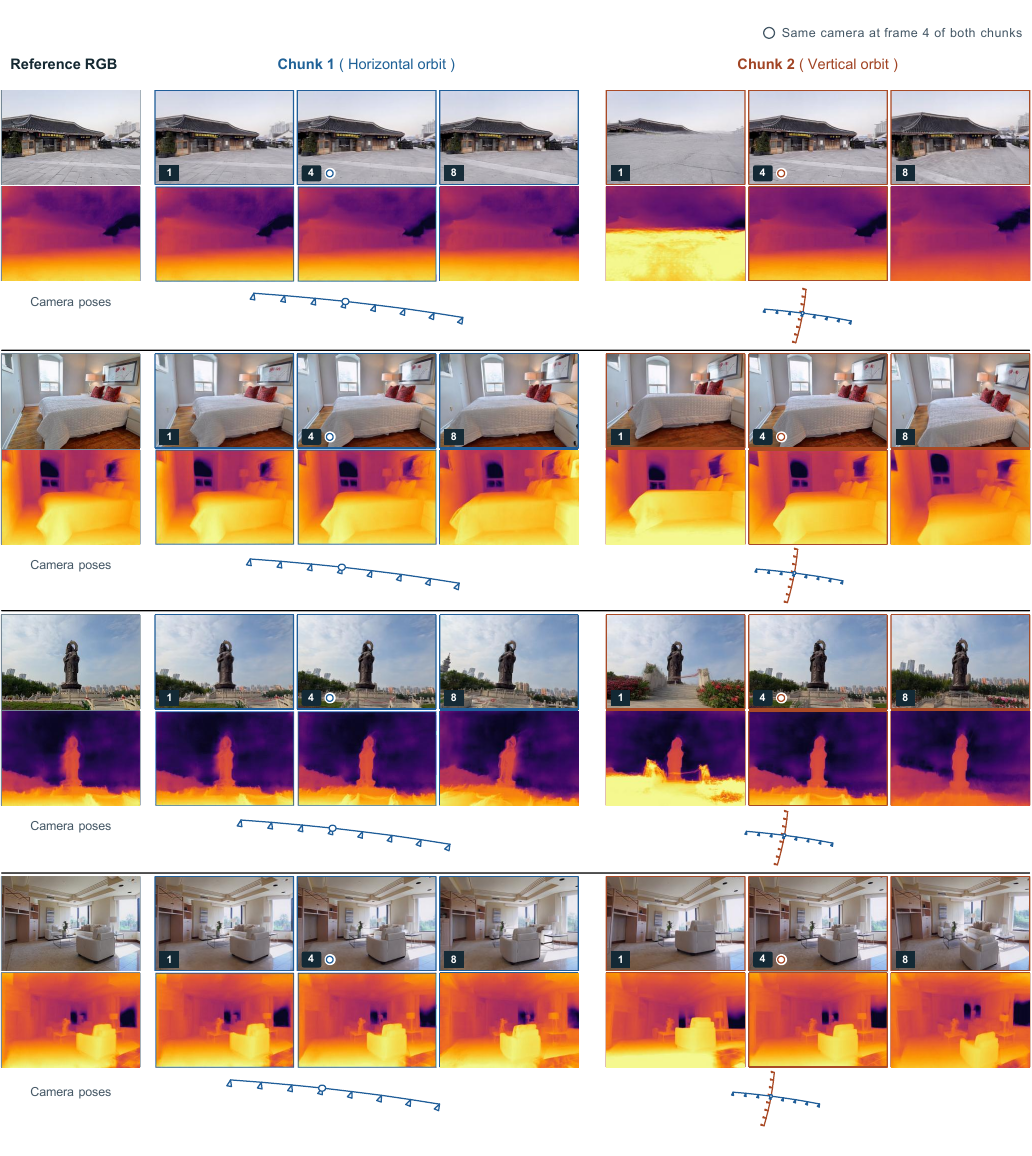}%
}\par\vfill

\newpage
\par\vspace*{\fill}%
\noindent\makebox[\textwidth][c]{%
    \includegraphics[pagebox=cropbox,trim=0bp 37bp 0bp 16bp,clip,width=\textwidth,
                     height=.9\textheight,keepaspectratio]{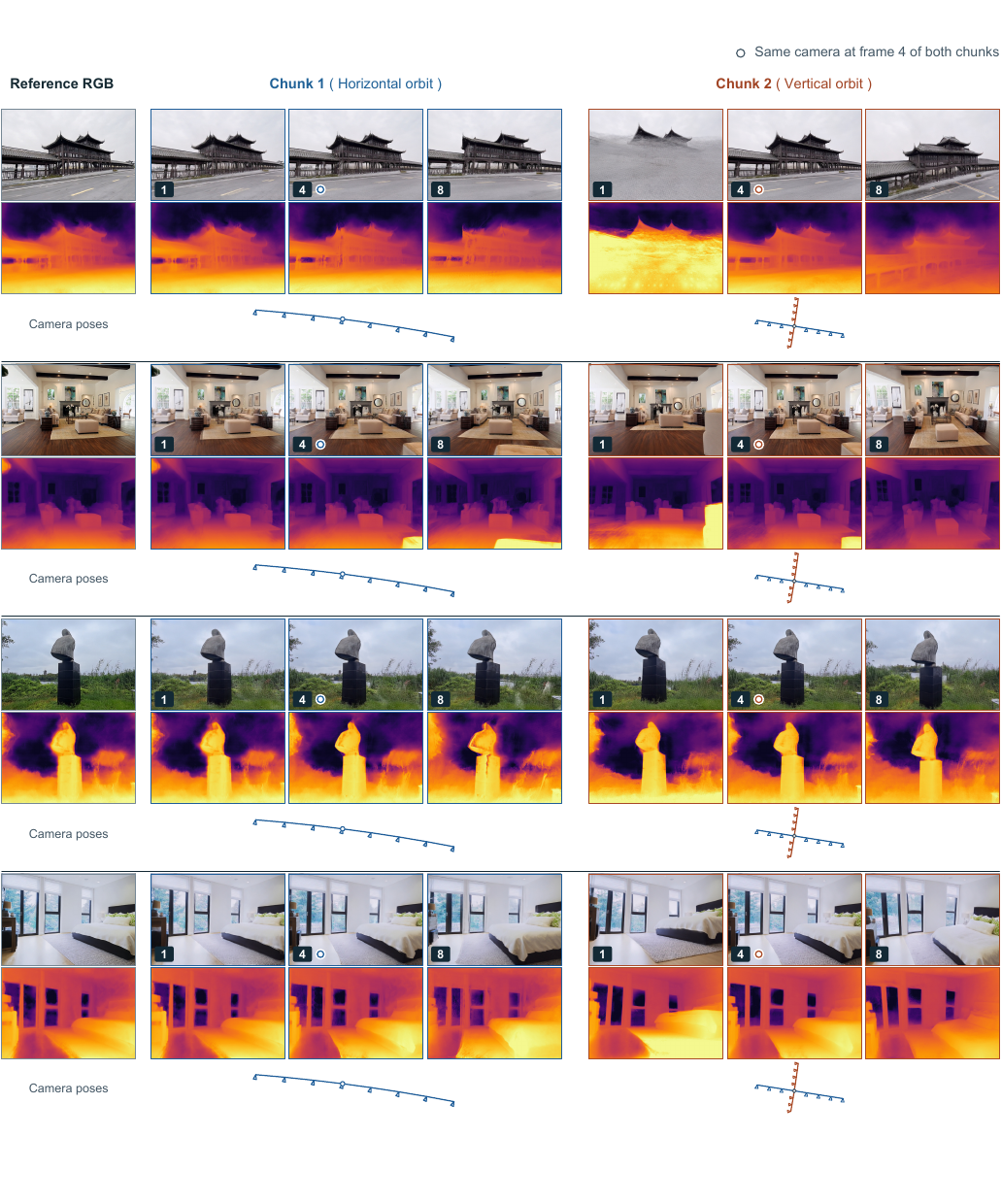}%
}\par\vfill

\end{document}